\documentclass[a4paper,14pt]{book}
\usepackage[utf8]{inputenc}

\usepackage[
backend=biber,
style=numeric,
sorting=none
]{biblatex}
\usepackage{graphicx}
\usepackage{float}
\usepackage[section]{placeins}
\usepackage[toc,page]{appendix}
\usepackage{setspace}
\usepackage{hyperref}
\usepackage{longtable}
\usepackage{booktabs}
\usepackage{caption} 
\usepackage{dirtytalk}
\usepackage{lscape}

\begin{document}



\thispagestyle{empty}
\vspace*{1cm}
\hrule
\mbox{ }

\begin{center}
\begin{LARGE}
Answer ranking in Community Question Answering: a deep learning approach \\ [0.3em]
\end{LARGE}
\end{center}

\hrule
\vspace*{1cm}
\begin{center}
 \includegraphics[width=0.3\textwidth]{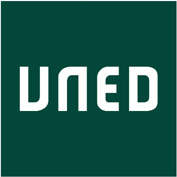} 
\end{center}

\vfill
\begin{center}
  {\Large {\bf Trabajo Fin de Máster}}
\end{center}

\vfill
\begin{center}
  \textbf{Lucas Valentin Garcia} \\ [0.2em]
  Trabajo de investigación para el \\ \vspace{10pt}
  Máster en Tecnologías del lenguaje\\ \vspace{10pt}
  Universidad Nacional de Educación a Distancia\\ \vspace{10pt}
  \vspace{2cm}
  Dirigido por  \\ \vspace{10pt}
  \textbf{Prof. Dr. D. Álvaro Rodrigo} \\ [0.5em]
  \textbf{Prof. Dr. D. Roberto Centeno} \\ [0.5em]
  Septiembre 2022\\
\end{center}

\chapter*{Abstract}
\label{sec:Abstract}
\noindent Community Question Answering is the field of computational linguistics that deals with problems derived from the questions and answers posted to websites such as Quora or Stack Overflow. Among some of these problems we find the issue of ranking the multiple answers posted in reply to each question by how informative they are in the attempt to solve the original question.

\noindent This work tries to advance the state of the art on answer ranking for community Question Answering by proceeding with a deep learning approach. We started off by creating a large data set of questions and answers posted to the Stack Overflow website.

\noindent We then leveraged the natural language processing capabilities of dense embeddings and LSTM networks to produce a prediction for the accepted answer attribute, and present the answers in a ranked form ordered by how likely they are to be marked as accepted by the question asker.

\noindent We also produced a set of numerical features to assist with the answer ranking task. These numerical features were either extracted from metadata found in the Stack Overflow posts or derived from the questions and answers texts.

\noindent We compared the performance of our deep learning models against a set of forest and boosted trees ensemble methods and found that our models could not improve the best baseline results. We speculate that this lack of performance improvement versus the baseline models may be caused by the large number of out of vocabulary words present in the programming code snippets found in the questions and answers text.

\noindent We conclude that while a deep learning approach may be helpful in answer ranking problems new methods should be developed to assist with the large number of out of vocabulary words present in the programming code snippets.

\tableofcontents
\listoffigures
\listoftables

\chapter{Introduction}

Community Question Answering \cite{srba} is the field of computational linguistics that relates to the analysis of questions and answers posted by users of community question answering websites such as Quora\cite{quora} or Stack Overflow\cite{stackoverflow}. In these websites users may freely post questions either in an unrestricted domain (Quora) or in sites specific domains (Stack Overflow). Other website users will then freely post answers that aim to answer the original question, making use of their personal experience and knowledge, or other Internet resources, such as Wikipedia or topic specific websites or manuals.

This simple mechanic introduces several challenges to both site users and moderators or maintainers. First of all the huge volume of posts (either questions or answers to these questions) requires some automated way to organize, categorize and quantify all this information. Some of the most common problems found in community question answering sites related to the questions deal with detecting duplicated questions that were answered previously or detecting questions that do not have a definite answer as they lend themselves to a debate among peers rather than a unique answer. On the array of problems related to the answers we may find the need to select the best answers, either by highlighting a single answer or sorting all the possible answers in a ranked way. Also the answers may be classified depending on whether they are informative or they simply raise a point in a debate but do not offer a conclusive answer.

Given that all these problems require sifting through large amounts of information it is immediate to apply techniques and procedures from the machine learning and natural language processing fields to attack some of these problems.

In this work we will propose a method to rank multiple answers to questions in community question answering sites found in a data set compiled from Stack Overflow postings. This ranking will be performed using deep learning models fed with dense embeddings of textual data, in addition to numerical features extracted both automatically and manually from the data set.

The reasoning behind these methods is that we believe that it is possible to augment the answer ranking capabilities of a deep learning model accepting a sequence of tokens as input with numerical features derived from the metadata relative to users and posts. In some cases these numerical features will be obtained immediately from the metadata while other numerical features will be created after parsing these metadata and the questions and answers texts.

\section{Objectives}

This MsC dissertation aims to advance the state of the art on community Question Answering (cQA). More specifically we will attempt to delve into new methods to predict the answer selected by the user who posted the question. The reasons to select a concrete answer may be either attending to the usefulness of the answer, how factual it is or even how soon it was posted. In order to better represent the results of the selected answer prediction we will produce a ranking of the multiple answers per question, sorted by how confident the model is that a given answer will be selected as accepted by the user.

Among the more specific objectives we will tackle we could enumerate:

\begin{itemize}
  \item Produce a relevant data set composed of information retrieved from community Question Answering sites. This data set will be composed by questions and answers text at a minimum, and may be incremented with additional metadata from either the user asking the question or the user answering the question.
  \item Analyze the data set and extract additional predictive features in addition to the existing metadata. These predictive features should increment the capabilities of the prediction model in order to improve the prediction results versus a simpler model that only makes use of the textual input.
  \item Identify sensible metrics that allow a realistic evaluation of the performance reached by the prediction model. Ideally these metrics should point out when a model is not able to rank the accepted answer in first place but is still able to place the answer in a relevant position so it can be clearly distinguished from the rest of answers.
  \item Carry out different experiments that allow us to gauge how each of the different sections of the models contribute to their performance. We will experiment with different options for the contextual embedding, number of recurrent and predictive layers, and also whether considering only the answer or the ensemble of the question and the answer effect the model performance.
  
\end{itemize}

\section{Structure}
This dissertation is organized as follows: 

\begin{itemize}
  \item On Chapter 2 we will first review the current state of the art on the community question answering, along with some of the most relevant problems in this field.
  \item On Chapter 3 we will present the data set used in our research, describing the process to obtain it and then describing the features in the data set. We will also introduce the methodology to evaluate the experimental results we obtain. Finally we will introduce a set of sensible baseline models that provide a minimal performance level against which we will later compare our models.
  \item On Chapter 4 we will describe the experiments we propose, including different ablation studies.
  \item On Chapter 5 we will review the results we obtain from the experiments, both from the initial baselines and from the more advanced models we created.
  \item On Chapter 6 we will discuss the results presented before and the relations between the baseline results and the advanced models results.
  \item Finally on Chapter 7 we will draw some conclusions from the work presented here and introduce some future line of work related to the CQA task.
\end{itemize}

\chapter{State of the art}

\section{Introduction}

In this section we will describe the current state of the art for community question answering. We will first describe the differences between classical question answering and community question answering problems. Then we will review the current approaches to community question answering and introduce the answer ranking problem and how it has been tackled in the past. Finally we will review some of the novel data sets related to community question answering.

\section{Community Question Answering}

Community Question Answering is the field of computational linguistics that deals with the problems arising from questions and answers posted by users to websites such as Quora or Stack Overflow \cite{10.1145/2009916.2009974}.

It is a field with a relatively long history, since the online communities where users may post questions to be answered by other users have now existed for many years. 

Community questions answering must not be confused with the classical question answering field where machine learning methods are used to obtain answers to questions about a given corpus. Question answering problems have a much more close ended nature since it is expected that the question may be fully answered with information present in the input corpus, while community question answering problems are much more open ended in the sense that questions posted by the users may refer a plurality of fields and may require external sources of information to be  properly answered.

\section{Community Question Answering problems}

We can find many examples of different research topics, all of them inside the wide field of CQA
(\cite{srba}, \cite{Roy}).

Some of the most common problems in CQA are related to the questions contents, such as duplicate question detection, question answerability prediction or quality assessment. Other classical CQA problems relate to the answers quality, such as the answer quality assessment, best answer prediction and, finally, answer ranking.

We will now see each of these problems in more detail. First duplicate question detection focuses on identifying existing questions to which newly posted question most resemble so that question answerers don't spend time on these questions. For instance Xu et al. \cite{10.1145/3239235.3240503} showed that a simple SVM approach may work better than deep learning methods for related question prediction.

The problem of question answerability prediction deals with the fact that some questions remain unanswered even years after they have been posted. This may be due to bad wording on the question or the fact that the question is indeed difficult to be answered, especially by people that do not have the questioner's experience or background. Asaduzzaman et al. \cite{6624015} found that the number of unanswered questions has increased significantly in the previous 2 years. Then they proposed a classifier that predicts how long a question will remain unanswered that reaches precision values of 0.38 and recall values of 0.45.

Nowadays, CQA sites are experiencing a surge in subpar content. The task of question quality assessment tries to identify high value questions for which it is more worth to spend time answering them. Maxwell Harper et al. \cite{harper}  created a model that reaches 89.7\% classification accuracy identifying informative questions from those only posted for conversational reasons.

Many other problems in CQA relate to answers rather than questions. For instance the problem of answer quality assessment refers to the methods proposed to measure the content of individual answer quality, distinguishing between high quality and low quality answers that bloat the answers page. Suguu et al. \cite{suggu-etal-2016-hand} achieved a 75.2\% accuracy predicting high quality questions with a bidirectional LSTM network with feature fusion model. 

Another problem related to answers in CQA is the best answer prediction, as selected by the user posting the original question. Gkotsis et al. \cite{10.1145/2615569.2615681} used a fusion of textual features with user and answer ratings features to achieve an 84\% average precision and 70 \% recall. It is to note that they reached this performance thanks to the discretization of their features so they are evaluated in the context of all the possible answers to the questions.

\section{Answer ranking in Community Question Answering}

There have been different attempts at ranking answers to questions in community question answering sites. Some of these attempts followed classical approaches like \cite{10.1145/2065023.2065030} which used similarity measure between questions and answers to rank the different answers. The work leveraged snippets of web search results for query expansion in answer ranking.

Ginsca et al. \cite{10.1145/2513577.2513579} used in-depth analysis of the information provided by the users in their profiles in order to discriminate features that are correlated to expertise, focusing then on profile and activity related features to rank answers.

Dalip et al. \cite{10.1145/2484028.2484072} followed a learning to rank approach based on different groups of features like features referred to the users, stylistic or structural features.

Bougessa et al. \cite{10.1145/1401890.1401994} aimed to identify expert users, on the basis that their answers have more relevance than answers posted by other users. The authority scores of users were modeled with a mixture of gamma distributions.

Amancio et al. \cite{AMANCIO2021102552} used recency and quality as criteria to rank answers, on the grounds that a recent and high quality answer is preferable to a high quality, date answer since users value more highly answers with more current content. The work considered an answer as recent not by how new is the date of
creation or editing of a given answer, but how current is the content of the answer .

More recently neural network approaches made possible by the advent of high performance GPUs have proved very valuable in the context of answer ranking in community question answering.

Zhou et al. \cite{ZHOU20188} used a gated recurent unit (GRU) \cite{https://doi.org/10.48550/arxiv.1412.3555} with thread-level features to rank the answers. The question and their answers were combined to create question-answer pairs, which are used as input to the model to capture the semantic features and then fed into the model to rank them.

Chen et al. \cite{10.1145/3077136.3080699} introduced a positional attention based recurrent neural network (RNN) model, which incorporates the positional context of the question words into the answers’ attentive representations, assuming that if a word in the question text occurs in an answer sentence, the neighboring words should be given more attention since they intuitively contain more valuable information.

Following the recent trend in many natural language problems, modern language models based on transformer with attention mechanisms that take an embedded representation input like BERT \cite{https://doi.org/10.48550/arxiv.1810.04805} have also been applied to the answer ranking task in community question answering.

Laskat et al. \cite{laskar-etal-2020-contextualized} used ElMo and BERT models both from a fine tuning approach as well as from a feature extraction approach to achieve state of the art results in the Semeval 2016 data sets.

Du et al. \cite{10.1007/978-3-030-93049-3_21} used Keyword BERT, a variation to the BERT model that adds a keyword-attentive layer that highlights the domain keywords to enhance the semantic interaction of the sentence pair supplied during training, in order to rank answers in a Buddhism related data set.

Maia et al. \cite{10.1007/978-3-030-86475-0_13} combined embedded representations of question and answer data with user defined tags for each question to form the input to the transformer model. In this way the model could derive the specific question domain for better answer ranking.

Wang et al. \cite{agriculture12020176} used a Chinese BERT model to extract embedded representations of question and answer text of a Chinese community answering site. Then, two matching strategies (Full matching and Attentive matching) were introduced in the matching layer to complete the interaction between sentence vectors. Finally a Bi-GRU network combined question and answer representations to obtain a measure of similarity to perform answer ranking.

\section{Community Question Answering data sets}

It is to note some of the latest efforts in the community question answering have focused on the curation of data sets that may enable new research lines and works. This trend started with the Yahoo! Answers dataset \cite{10.1145/1835449.1835518} which combined content from community question answering with manually generated annotations to obtain a dataset suitable for machine learning processing. 

The the Community Question Answering task in some of the past SemEval editions (\cite{https://doi.org/10.48550/arxiv.1911.11403, nakov-etal-2016-semeval-2016, nakov-etal-2017-semeval}) introduced a new data set retrieved from the Qatar Living forum \footnote{https://www.qatarliving.com/}, and proposed tasks related to question and answer similarity, and question to question similarity. On the 2016 and 2017 editions, the SemEval main task was to rank the answers to a question by how useful they were to answer the question. In addition, the 2017 edition introduced two new sub-tasks: rank the similar questions according to their similarity to the original question, and rank the answer posts according to their relevance with respect to the question. The metric used in these SemEval tasks is the mean average precision (MAP).

More recently the AmazonQA dataset \cite{https://doi.org/10.48550/arxiv.1908.04364} incorporated user generated reviews of products to the usual question and answer data set fields so new tasks such as automatic answer retrieval from a question review pair may be performed.

One common pitfall of all these data sets is that they mostly only collect question and answer text with few if any additional metadata. This complicates the task of producing additional non textual features that may be used in problems like answer ranking.

\section{Conclusions}

As we have seen there have been multiple attempts at answer ranking in community question answering. While it was common to use non textual features before the advent of large neural network models, we have not found many instances of modern models that make use of these non textual features. It may be fruitful then to combine modern neural network models with an ensemble of non textual features and verify whether the addition of these features improves the answer ranking capabilities.

In addition it is required to collect a new data set that combines a large corpus of question and answer pairs with addition data from users and posts that allow the extraction of informative features for answer ranking.
\chapter{Evaluation framework}

\section{Introduction}

In this chapter we will introduce the elements that will allow us to conduct experiments on the answer ranking problem. We will start by describing the data set that has been compiled from Stack Overflow posts in order to perform the answer ranking task over community question answering content. Next we will introduce different metrics that we will use to assess the performance of the proposed methods. Finally we will present a set of baseline models based around random forests and gradient boosted trees that make use of simple Bag of Words text representation.

\section{Data}

\subsection{Stack Overflow data}

We will make use of data from the largest community question answering site nowadays as is Stack Overflow. In addition the fact that data from such site is easily accessible through Google BigQuery \cite{hoffa} will greatly simplify the data collection task.

Stack Overflow is a community question answering created in 2008 and nowadays it has over 14 million registered users \cite{wiki:Stack_Overflow}. More than 21 million questions and 31 million answers have been posted to the site. Based on the type of tags assigned to questions, the top eight most discussed topics on the site are: JavaScript, Java, C\#, PHP, Android, Python, jQuery, and HTML. 

\begin{figure}[h]
    \centering
    \includegraphics[width=\textwidth,height=\textheight,keepaspectratio]{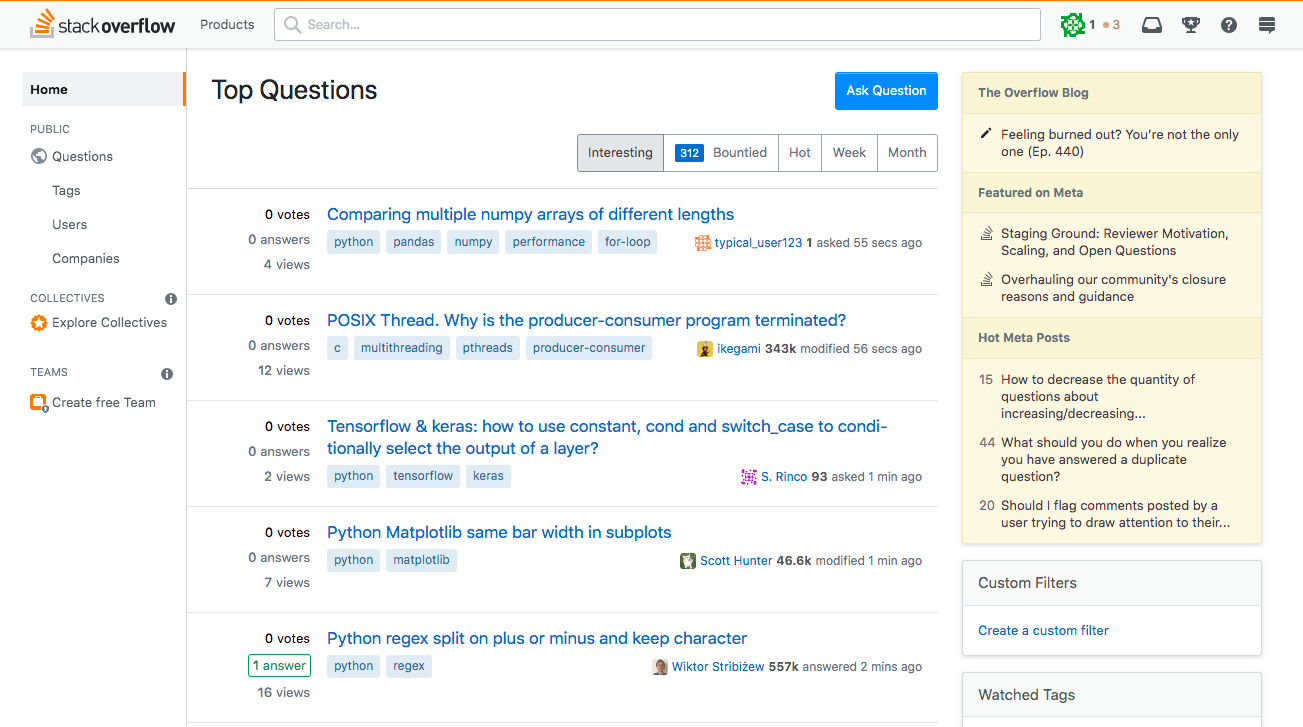}
    \caption{Capture of the Stack Overflow site home page}
    \label{fig:stackoverflow-home}
\end{figure}

\begin{figure}[h]
    \centering
    \includegraphics[width=\textwidth,height=\textheight,keepaspectratio]{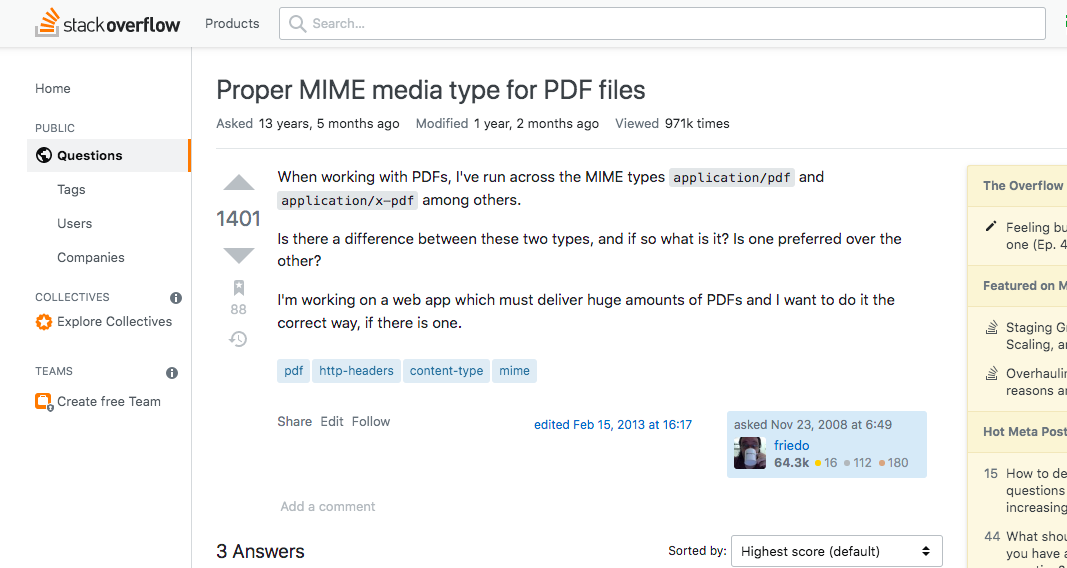}
    \caption{Capture of an example question on Stack Overflow}
    \label{fig:stackoverflow-q}
\end{figure}

\begin{figure}[h]
    \centering
    \includegraphics[width=\textwidth,height=\textheight,keepaspectratio]{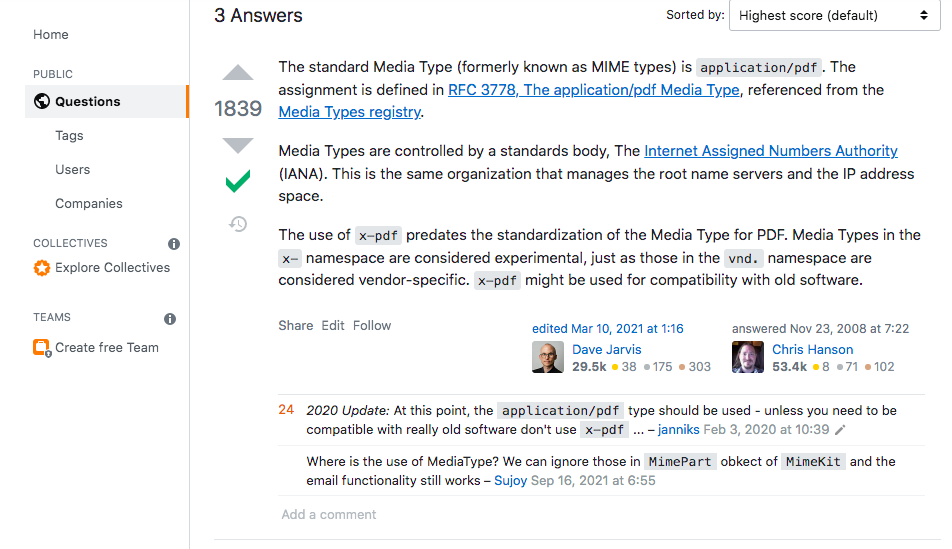}
    \caption{Capture of an example answer on Stack Overflow}
    \label{fig:stackoverflow-a}
\end{figure}

There are many data items available for Stack Overflow posts. We will review some of these data items, focusing on the ones most relative to our answer ranking task.

\begin{figure}[h]
    \centering
    \includegraphics[scale=0.5]{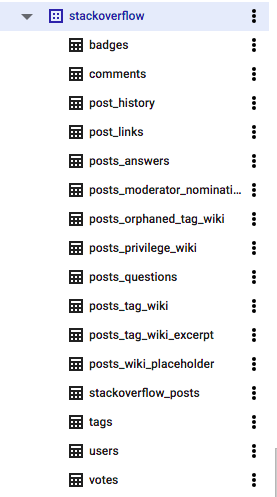}
    \caption{Capture from BigQuery site where all the tables with Stack Overflow data are shown }
    \label{fig:stackoverflow-tables}
\end{figure}

Given that the Stack Overflow site data we are accessing is available through the database in Google BigQuery we find that all data is available as database tables. More specifically, all the Stack Overflow data is scattered through different tables as we can see in figure \ref{fig:stackoverflow-tables}

We will now briefly describe each of these tables with a special focus on the ones holding items of interest for our task.

\begin{enumerate}
    \item badges: this table holds the badges collected by each user of the site either by posting questions or answers. This table will not be of interest for our task.
    \item comments: this table holds the comments posted by users to some of the answers. Usually these comments are posted by users who also replied to the original question so they may not be very informative but rather engage in punctilious analysis that may often miss the point of the original question. This table will not be of interest for our task.
    \item post\_history: this table holds details on when questions or answers where edited.  This table will not be of interest for our task.
    \item post\_links: this table holds details on which posts in the site link to other posts. This table will not be of interest for our task.
    \item posts\_answers: This table holds the answers posted to the questions in the site. This table if of great interest for our task so we will review it in detail, describing each of the columns this tables holds.
    \begin{enumerate}
        \item id: this column holds the answer ID. It is not directly applicable to our task but we will use it to link each answer to the question it refers to.
        \item title: this column holds no values so it will not be of interest for our task.
        \item body: this column holds the text of the answer. It will be very important for our task.
        \item accepted\_answer\_id: this column holds no values so it will not be of interest for our task.
        \item answer\_count: this column holds no values so it will not be of interest for our task.
        \item comment\_count: this column holds the number of comments posted to the answer. It may be important so we will use as a numerical feature in our experiments.
        \item community\_owned\_date: this column holds no values so it will not be of interest for our task.
        \item creation\_date: this columns holds the timestamp for the answer creation. We will not use it for our task.
        \item favorite\_count: this column holds no values so it will not be of interest for our task.
        \item last\_activity\_date: this column holds the timestamp for when the answer last saw any activity like posting or edits. We will not use it for our task.
        \item last\_edit\_date: this column holds the timestamp for when the answer was last edited. We will not use it for our task.
        \item last\_editor\_display\_name: this column holds no values so it will not be of interest for our task.
        \item last\_editor\_user\_id: this column holds the user ID that last edited the answer. We will not use it for our task.
        \item owner\_display\_name: this column holds no values so it will not be of interest for our task.
        \item owner\_user\_id: this column holds the user ID that posted the answer. We will not use it for our task.
        \item parent\_id: this columns holds the ID for the question it refers to.  It is not directly applicable to our task but we will use it to link each answer to the question it refers to.
        \item post\_type\_id: this column holds the type of post it refers to. All rows contain value 2 for answer hence the column has no informative value and we will not use it for our task.
        \item score: this column holds the score obtained by the answer after users up-vote and down-vote it, attending to its contents. It may be very important for our task so we will use it as a numerical feature.
        \item tags: this column holds no values so it will not be of interest for our task.
        \item view\_count: this column holds no values so it will not be of interest for our task.

    \end{enumerate}
    \item posts\_moderator\_nomination: this table holds the self nominations from users who would like to be promoted to moderators in the site. This table will not be of interest for our task.
    \item posts\_orphaned\_tag\_wiki: this table holds information from the wiki section of Stack Overflow. This table will not be of interest for our task. 
    \item posts\_privilege\_wiki: This table is also related to the wiki section of Stack Overflow and will not be of interest for our task.
    \item posts\_questions: this table holds the questions posted by users of the site. This table if of great interest for our task so we will review it in detail.
    \begin{enumerate}
        \item id: this column holds the question ID. It is not directly applicable to our task but we will use it to link each answer to the question it refers to.
        \item title: this column holds the question title, which introduces the topic of the question. It will be very important for our task.
        \item body: this column holds the question body, where the question is developed. It will be very important for our task.
        \item accepted\_answer\_id: this column holds the ID for the accepted answer. It is not directly applicable to our task but we will use it to assess whether each answer was marked as accepted or not.
        \item answer\_count: this column holds the number of answers posted to each question. It may be important so we will use as a numerical feature in our experiments.
        \item comment\_count: this column holds the number of answers posted to each question. It may be important so we will use as a numerical feature in our experiments.
        \item community\_owned\_date: this column holds no values so it will not be of interest for our task.
        \item creation\_date: this columns holds the timestamp for the question creation. We will not use it for our task.
        \item favorite\_count: this column holds the count for the number of times any user (who may or may not be the one who posted the question) set the question as favourite in order to track its status. We will not use it for our task.
        \item last\_activity\_date: this column holds the timestamp for when the answer last saw any activity like posting or edits. We will not use it for our task.
        \item last\_edit\_date: this column holds the timestamp for when the answer was last edited. We will not use it for our task.
        \item last\_editor\_display\_name: this column holds no values so it will not be of interest for our task.
        \item last\_editor\_user\_id: this column holds the user ID that last edited the answer. We will not use it for our task.
        \item owner\_display\_name: this column holds no values so it will not be of interest for our task.
        \item owner\_user\_id: this column holds the user ID that posted the answer. We will not use it for our task.
        \item parent\_id: this columns holds the ID for the answer it refers to.  It is not directly applicable to our task but we will use it to link each answer to the question it refers to.
        \item post\_type\_id: this column holds the type of post it refers to. All rows contain value 1 for questions hence the column has no informative value and we will not use it for our task.
        \item score: this column holds the score obtained by the question after site users up-vote and down-vote it, attending to its contents. It may be very important for our task so we will use it as a numerical feature.
        \item tags: this columns holds the tags sets by the user posting the question. These tags usually describe the domain to which the question belongs and could be used to separate questions into different domains. We will not be using for our experiments since we are not interested in discriminating questions by domain for now.
        \item view\_count: this column holds the count for the number of times the question was accessed by users on the site. We will not use it for our task.
    \end{enumerate}
    \item posts\_tag\_wiki: This table is related to the wiki section of the site and will not be of interest for our task.
    \item posts\_tag\_wiki\_excerpt: This table is related to the wiki section of the site and will not be of interest for our task.
    \item posts\_wiki\_placeholder: This table is related to the wiki section of the site and will not be of interest for our task.
    \item stackoverflow\_posts: this table is a legacy element that used to hold both questions, answers and comments. It is recommended by the data set documentation not to use it and use tables with names beginning with posts\_ instead.
    \item tags: this table holds the tags assigned by users to each question. These tags may be used to categorize the question into different domains. This table will not be of interest for our task.
    \item users: this table holds data on users posting questions and answers to the site. This table if of great interest for our task so we will review it in detail.
    \begin{enumerate}
        \item id: this column holds the question ID. It is not directly applicable to our task but we will use it to link each user to the answer  it posts.
        \item display\_name: this columns shows the pseudonym that the user choose to represent herself in the site. We will not use it for our task.
        \item age: this columns holds the user age. It may be important so we will use as a numerical feature in our experiments.
        \item creation\_date: this column holds the timestamp when the user created the account in the site. We will not use it for our task.
        \item last\_access\_date: this column holds the timestamp when the user last accessed the sire. We will not use it for this task.
        \item location: this columns holds the user location. It may be important so we will use as a numerical feature in our experiments.
        \item reputation: this column holds the user reputation reached as a consequence of posting questions and answers to the site. It may be important so we will use as a numerical feature in our experiments.
        \item up\_votes: this columns holds the number of up votes received by the user contributions. It may be important so we will use as a numerical feature in our experiments.
        \item down\_votes: this columns holds the number of down votes received by the user contributions. It may be important so we will use as a numerical feature in our experiments.
        \item views: this column holds the count of accesses to the user profile in the site. It may be important so we will use as a numerical feature in our experiments.
        \item profile\_image\_url: this column holds the URL address for the user profile picture. It may be important so we will use as a numerical feature in our experiments.
        \item website\_url:  this column holds the URL address for user website. It may be important so we will use as a numerical feature in our experiments.
    \end{enumerate}
    \item votes: this table holds data on votes to questions and answers casted by users. This table will not be of interest for our task.
\end{enumerate}

\subsection{Data retrieval}

As we have said we will retrieve Stack Overflow site data through the Google BigQuery service. BigQuery provides an Structured Query Language (SQL) interface so data may accessed through any computing environment that supports such interface. Given that as we have described in the previous section the data that we will be using is spread across several tables we will need to create SQL queries that join the data in these tables.

A detailed description of the SQL queries used to obtain our data would be beyond the scope of this dissertation but we will summarize some of the most important features of the SQL queries in this section. The full text to the SQL queries may be found at Appendix \ref{appendix:sql}.

First we will make use of the question and answer ID codes found in the question and answer tables to join each of the answers in the answers table with the respective questions being answered. In addition we will also make use of the user ID referring to the user who created each answer to join user data to the answer data.

Once we have joined the data from the questions and answers tables we will extract data from the relevant columns identified before to form our data set.

In addition to these join commands we will also perform some filtering in the data: we will only grab data from questions with more than one answer and from questions with an accepted answer. The goal of this filtering is to ensure that all questions in our data set have an accepted answer and more than one answer so ranking them is a non trivial issue.

For our experiments we will our models with balanced data respect to the number of accepted answers; this is, we will use data collections with the same number of accepted answers as of not accepted answers.

Finally we will filter posts with questions asked in 2016 for the training set and posts with questions asked in 2017 for the test test.

\subsection{Additional Feature Generation}

After performing data retrieval from BigQuery we will produce extra features derived from this retrieved data. In particular we will produce some features relative to the question and answer texts such as the word count (n\_words), the average word length (avg\_word\_len), the sentence count (n\_sent) and the average and maximum number of words per sentence (avg\_n\_word\_sent, max\_n\_word\_sent). In addition we will also produce an additional feature that will count the number of common words between question and answer (qa\_n\_common), and an additional feature that will inform whether there were HTTP hyperlinks in the answer body (a\_has\_urls).

Also for some of the columns in the retrieved data we will encode the presence of absence of data as a Boolean variable, given that the exact content of these columns is not informative but rather the presence or absence is. More specifically we will create Boolean variables to flag the presence of information about the user such as the "About" field of the user's profile, the user's location, the existence of a profile image or a website in the user's profile (has\_user\_about, has\_user\_location, has\_user\_profile\_image\_url, has\_user\_website\_url).

\subsection{Exploratory Data Analysis}

In this section we will examine closely the features produced in the previous section.

\begin{figure}[!htb]
\caption{a\_accepted class distribution}
\label{img:eda-class-dist}
\centering
\includegraphics[width=0.5\textwidth]{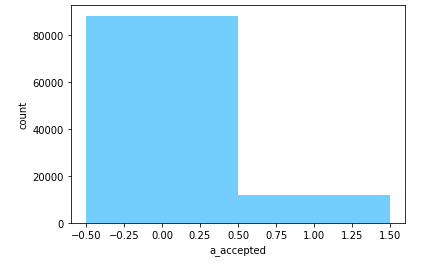}
\end{figure}

First of all we will check the distribution of a\_accepted as this is the target of our answer selection task, as can be seen in figure \ref{img:eda-class-dist}.

\begin{figure}[!htb]
\caption{a\_accepted feature correlation heat map}
\label{img:eda-heat-map}
\centering
\includegraphics[width=1.0\textwidth]{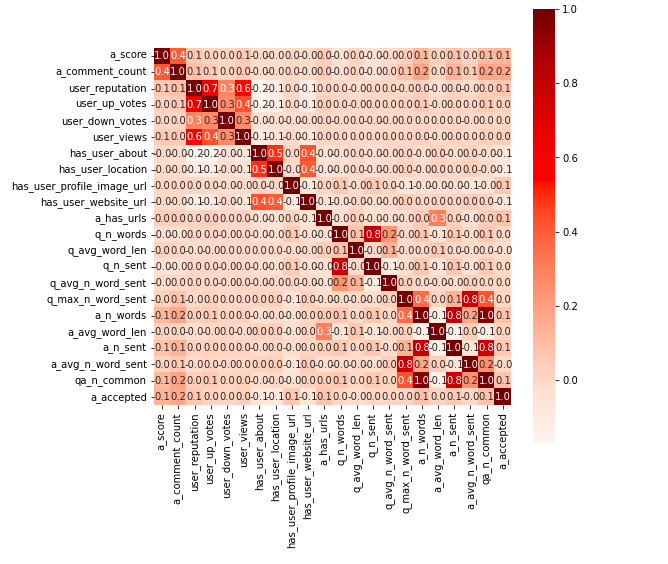}
\end{figure}

As we can see there is a great imbalance as there are many more not accepted answers that accepted answers. This is something perfectly logical since by definition only one answer per question will be marked as accepted. Now we will produce a heat map of the different features correlation to the a\_accepted class, as can be seen in figure \ref{img:eda-heat-map}.

As we can see in the heat map for all of the features the correlation value is rather low. We will see it better in table \ref{table:1}, where we observe that just a few of the features have a correlation to class higher than 0.1.

\begin{table}[h]
\centering
\begin{tabular}{lr}
                   Feature &  a\_accepted correlation \\
         has\_user\_location &               -0.089347 \\
            has\_user\_about &               -0.077484 \\
      has\_user\_website\_url &               -0.068921 \\
            q\_avg\_word\_len &               -0.017228 \\
         a\_avg\_n\_word\_sent &               -0.000275 \\
                  q\_n\_sent &                0.002184 \\
                 q\_n\_words &                0.002589 \\
           user\_down\_votes &                0.008166 \\
         q\_max\_n\_word\_sent &                0.011173 \\
         q\_avg\_n\_word\_sent &                0.014262 \\
            a\_avg\_word\_len &                0.018356 \\
             user\_up\_votes &                0.023303 \\
                user\_views &                0.032269 \\
                  a\_n\_sent &                0.056832 \\
has\_user\_profile\_image\_url &                0.058864 \\
           user\_reputation &                0.066050 \\
                a\_has\_urls &                0.072321 \\
                 a\_n\_words &                0.076058 \\
               qa\_n\_common &                0.080131 \\
                   a\_score &                0.128801 \\
           a\_comment\_count &                0.168658 \\
\end{tabular}
\caption{a\_accepted class to feature correlations}
\label{table:1}
\end{table}

Given that most of the features (either retrieved in the SQL query or generated manually) have an extremely low correlation to the a\_accepted class we can presume that they will not be very useful to predict the selected answer.

Due to the low correlation of most features we will continue our exploratory data analysis only with those features where we observe an absolute correlation value with the a\_accepted class higher than 5\%. As we can see in table \ref{table:2} these variables will be has\_user\_location, has\_user\_about, has\_user\_website\_url, a\_n\_sent, has\_user\_profile\_image\_url, user\_reputation, a\_has\_urls, a\_n\_words, qa\_n\_common, a\_score and a\_comment\_count.

\begin{table}[!htb]
\centering
\begin{tabular}{lr}
                   Feature &  a\_accepted correlation \\
         has\_user\_location &               -0.089347 \\
            has\_user\_about &               -0.077484 \\
      has\_user\_website\_url &               -0.068921 \\
                  a\_n\_sent &                0.056832 \\
has\_user\_profile\_image\_url &                0.058864 \\
           user\_reputation &                0.066050 \\
                a\_has\_urls &                0.072321 \\
                 a\_n\_words &                0.076058 \\
               qa\_n\_common &                0.080131 \\
                   a\_score &                0.128801 \\
           a\_comment\_count &                0.168658 \\
\end{tabular}
\caption{Selected a\_accepted class to feature correlations}
\label{table:2}
\end{table}

\begin{figure}[!htb]
\caption{Selected features histogram split per class}
\label{img:eda-hist-all}
\centering
\includegraphics[width=1.0\textwidth]{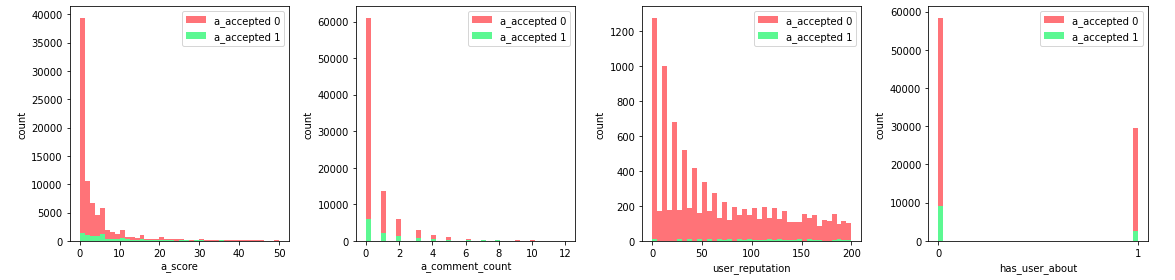}
\includegraphics[width=1.0\textwidth]{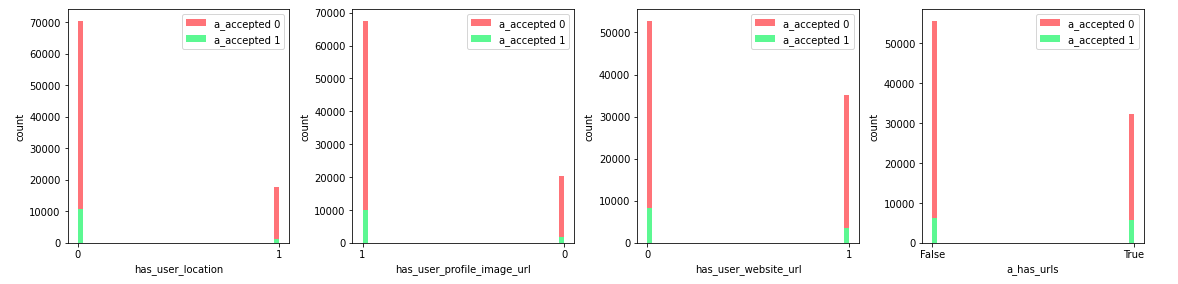}
\includegraphics[width=0.75\textwidth]{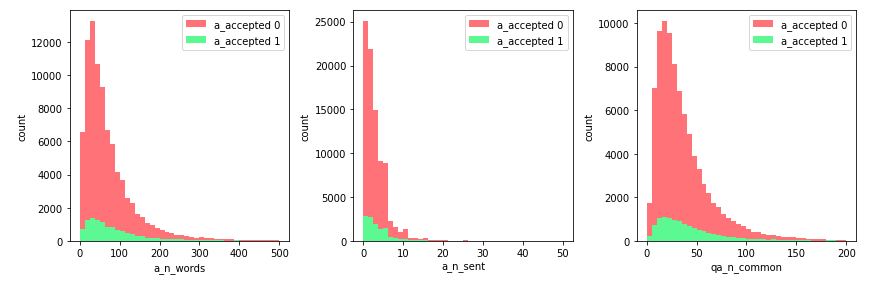}
\end{figure}

We will now explore the distribution for these features with respect to the a\_accepted class. In order to do so we will produce individual histogram plots per feature, splitting the histogram samples per class. Figure \ref{img:eda-hist-all} show these histograms. As we can see, due to the low correlation between each of these features and the a\_accepted class there is not a large difference between the distribution of values for each feature across the class.

\section{Metrics}

As we have seen when the data set was described the original class distribution for the accepted answer feature is very unbalanced, hence using a classical binary classification metric such as accuracy would be misleading since a naive model that predicted all answers to not be accepted would yield a large accuracy values.

Due to this class unbalancing it was decided to use figures of merit from the information retrieval field, where multiple results are available for a single query and different metrics allow to easily compare between the different results which ones have been better classified or ranked. The information retrieval metrics we will use are the Mean Reciprocal Rank (MRR) \cite{voorhees-tice-2000-trec} and the Normalized Discounted Cumulative Gain (NDCG) \cite{10.1145/582415.582418}. 

It was decided to discard the usage of Mean Average Precision (MAP) metric which is the most common in information retrieval problems as we only consider one correct answer per question: the answer accepted by the user who posted the question. In this case MAP will equal the MRR values, so there is no need to add MAP to the metric suite.

In order to define the Mean Reciprocal Rank we will first define the Reciprocal Rank. The Reciprocal Rank measure calculates the reciprocal of the rank at which the first relevant document resulting from an information retrieval query was retrieved. In our case this first relevant document will be the answer accepted by the user who asked the question. Given that we will be evaluating our experiments over a collection of questions and answers sets we will compute the average Reciprocal Rank over all the questions and corresponding answers sets, hence yielding the Mean Reciprocal Rank figure.

\[RR = \sum_{i=1}^{M} \frac{1}{rank_i} \]
\captionof{figure}{Reciprocal Rank equation}

\[MRR = \frac{1}{N} \sum_{i=1}^{N} RR_i \]
\captionof{figure}{Mean Reciprocal Rank equation}

The Normalized Discounted Cumulative Gain is another metric from the information retrieval field that measures the performance of an information retrieval system (most frequently a search engine) by ranking the results according to the their relevance in terms of the search query. In order to compute the Normalized Discounted Cumulative Gain we first must calculate the 
Discounted Cumulative Gain (DCG), this is the sum of the relevance scores of each of the query results divided by a discount factor such that lower ranking results have a smaller effect on the cumulative gain sum.

\[DCG_k = \sum_{i=1}^{k} \frac{relevance_i}{log_2 (i+1)} \]
\captionof{figure}{Discounted Cumulative Gain equation}

Then we will compute the Normalized Discounted Cumulative Gain by dividing the Discounted Cumulative Gain by the Ideal Discounted Cumulative Gain (IDCG), this is the DCG value for a perfect query, where each of the results has the actual importance.

\[NDCG_k = \frac{DCG_k}{IDCG_k}\]
\captionof{figure}{Normalized Discounted Cumulative Gain equation}

In the specific case of our experiments we will consider that the ideal relevance scores will take the form of a unity value for the accepted answer and zero for the rest of the answers.

We will see some examples of how these metrics will allow us to measure how well our models perform. Let us consider two arrays of values, one corresponding to the ground truth for the accepted answer and one for the model output. For the ground truth the array will resemble a one hot encoded vector, where only the index relative to the accepted answer is set to one will the other elements are set to zero. On the other hand the array corresponding to the model output will contain a graduation of values belonging to the probability that each of the answers to a question is the accepted answer, as predicted by the model under evaluation.

We will first consider the case of a perfect prediction, where the correct answer was perfectly identified. Hence we will have 2 arrays of data, the first one $r_{gold}$ that contains the ground truth of the accepted answer and $r_{eval}$ with the model output. In the case of a perfect input the value of each array will be

\[r_{gold}=\{0,0,1\}\ \, r_{eval}=\{0,0,1\}\]

On this example MRR and NDCG respectively will be 1.0 and 1.0 for this input.

We will now consider the case for the worst case prediction error. In this case $r_{gold}$ again contains the ground truth of the accepted answer and $r_{eval}$ with the model output, where the accepted answer was not the correct one:

\[r_{gold}=\{0,0,1\}\ \, r_{eval}=\{0,1,0\}\]

In this case MRR and NDCG respectively will be 0.333 and 0.565.

\section{Baselines}

We will now introduce a suite of sensible baseline models that allow us to set a minimum expectations for the methods we will propose later.

Given that the data we have collected is very heterogeneous in its nature due to the fact that it contains numerical and textual data belonging to both questions, answers and users we will need to create baseline models that reflect this plurality of data features.

Hence makes sense to split our baseline models into models that make use of only numerical data, models that make use of textual data and models that use a mixture of numerical and textual data.

As numerical data we will use the features described earlier, both the ones directly extracted from the BigQuery databse queries and the numerical features derived from the question and answer texts.

In the case of textual data we will produce vector space representations of the answer data by using TF-IDF (term frequency–inverse document frequency) weighting of the bag of words, after the usual pre-processing steps of HTML tag and punctuation symbol removal and lower case transformation.

We will pair this hierarchy of models with different forest models, namely random forests, gradient boosted trees and AdaBoost. We will only make use of forest based trees due to their well known features of properly handling large collections of tabular data and their immunity to over-fitting \cite{hastie01statisticallearning}.

One important thing to note is that, while the baselines we have described are all of them classification models with a discrete class output we will require to produce an output from them that may be used to rank the different answers by how likely they are to be the accepted answer. We will do so by getting the probability estimate output from each of the models, hence we will get a continuous, numerical figure that will easily allow us to rank each of the answers.

\subsubsection{Random Forests}

A random forest is an ensemble of decision trees which results in a better estimator than individual trees which suffer of over-fitting due to their low bias and high variance. By bagging a number of uncorrelated trees each of the high variance contributions of the individual trees are cancelled out. Other advantages of forests is their high parallelism and reduced complexity for training.

As their main disadvantage hand random forests lose the interpretability inherent to individual trees. Also, on problems where predictive features are very linear random forests may not improve the accuracy of individual trees by much.

\subsubsection{Gradient Boosted Trees}

Gradient boosting allows the improvement of predictions done by an ensemble of weak learners, by incrementally refining the predictive capabilities of its learners. Usually these weak learners take the form of a decision tree and the resulting model is called a gradient boosted trees model. At each step of boosting the ensemble of learners is modified to correct for the previous step prediction errors, effectively trying to minimize the error function between the ground truth and the ensemble predictions. The direction to take in order to update the learners is derived from the gradient of the error function following its opposite direction, hence the name gradient boosting.

Again the main disadvantage with respect to a decision tree is the loss of explainability due to the ensemble nature of the model.

\subsubsection{AdaBoost}

AdaBoost (short form for Adaptive Boosting) is another ensemble method that tries to overcome individual learners weaknesses in a serial fashion. As learner a simple tree with just 2 leaves is used, and each of the learners is assigned a weight updated by the model in each boosting round. This is different form random forests and gradient boosted trees where all the learners have equal weights.
\chapter{Proposed Methods}

\section{Introduction}

In this section we will introduce a suite of deep learning models that take embedded representations of the text contained in questions and answers as inputs, and additionally we will augment the model inputs with numerical features obtained from the data set. We will also introduce different ablation studies to observe the effect that different model parameters have on the model performance.

\section{Hypothesis}

We will enumerate a series of hypothesis that we will later test with the proposed experiments.

Firstly we propose that a Siamese neural network structure will allow us to infer informative relations between questions and answers text such that a meaningful performance level may be reached in terms of answer ranking.

Then we believe that augmenting the textual information from questions and answers with numerical features, either relative to the users or to the structure of the text like number of words per sentence, will allow our models to perform better.

Given that we will be using models that receive a sequence of tokens inputs we hypothesize that the maximum sequence length constraint may limit the model's performance.

Relative to these sequences of tokens we hypothesize that the corpus from which the embedding matrix is composed may affect the model performance, in the sense that when corpus that are closer to the online forum environments are used better model performance may be reached.

Finally we hypothesize that increasing the model complexity by adding more layers will improve the model performance for answer ranking.

\section{Models}

In this section we will describe the deep learning models we will use to attempt to improve the baseline results.

As with the baseline models we will also pair a plurality of models with different predictive features, like purely numerical features extracted from the data set, textual features from the question and answers text, and a mixture of both. In the case of numerical features we will simply use the same numerical data as in the baseline models. In the case of textual data we will produce dense embeddings using GloVe \cite{pennington2014glove} to represent the questions and answers in a high dimensional space so they are suitable to be used as inputs to the deep learning models.

We will create token sequences of dimension 100 using GloVe embeddings obtained from Wikipedia \cite{pennington2014glove}. The token sequences length will be limited to 100 tokens (both for question and answer texts), and we will be summarizing each question or answer text into two sentences by ranking the sentences by length and taking the top two. Among the many importance metrics we could use we will aim for low complexity and use a simple sentence length ranking, following the rationale that more meaningful sentences will be longer, due to the larger number of ideas they convey. In addition we will be balancing the training data set to contain 500 pairs of questions and accepted answers, and 500 pairs of questions and not accepted answers. The test set will be composed by 1000 pairs of questions and answers, following the same class imbalance seen in the exploratory data analysis performed earlier.

The first layer in all of our models will be a Long Short-Term Memory (LSTM) layer. LSTM layers are an especial case of Recurrent Neural Networks (RNNs), this is, neural networks where some of their internal state values are fed back into the input. This allows modelling temporal relations on an input sequence such as a sentence or series of sentences. In particular LSTMs allow overcoming the common issue seen in RNNs of vanishing gradient, seen when training artificial networks with gradient based methods and back-propagation. In these cases each of the network weights are updated by a factor directly proportional to the gradient of the error function with respect to the current weight. After a few training epochs some of these gradients may become very small, effectively preventing the weights from being updated and consequently stopping the network from training further. The existence of recurrent gates in the LSTM cell allows the errors to flow back hence preventing the vanishing gradient problem.

In our models we will make use of a bidirectional LSTM (Bi-LSTM), which uses a concatenation of two instances of the same replicated LSTM cell, with one of the instances seeing the sequence input from left to right and with the other instance seeing the sequence input from right to left. This allows the combined cell to process long term dependencies in either direction which is a crucial feature of natural language.

Since the textual data to be processed contains pairs of questions and answers we will replicate the input structure with a stack of Bi-LSTM cells twice, setting these 2 input structures branches in parallel so they receive question and answer text respectively. The output from the last layer in each of the parallel branches will be concatenated and fed into a dense layer that will act as a classification layer, producing a numerical output that will represent the probability that the input answer was selected as accepted by the question author.

\begin{figure}
\caption{Tensorflow representation for the deep learning model (including numerical features)}
\label{img:dl-base-model}
\centering
\includegraphics[width=\textwidth]{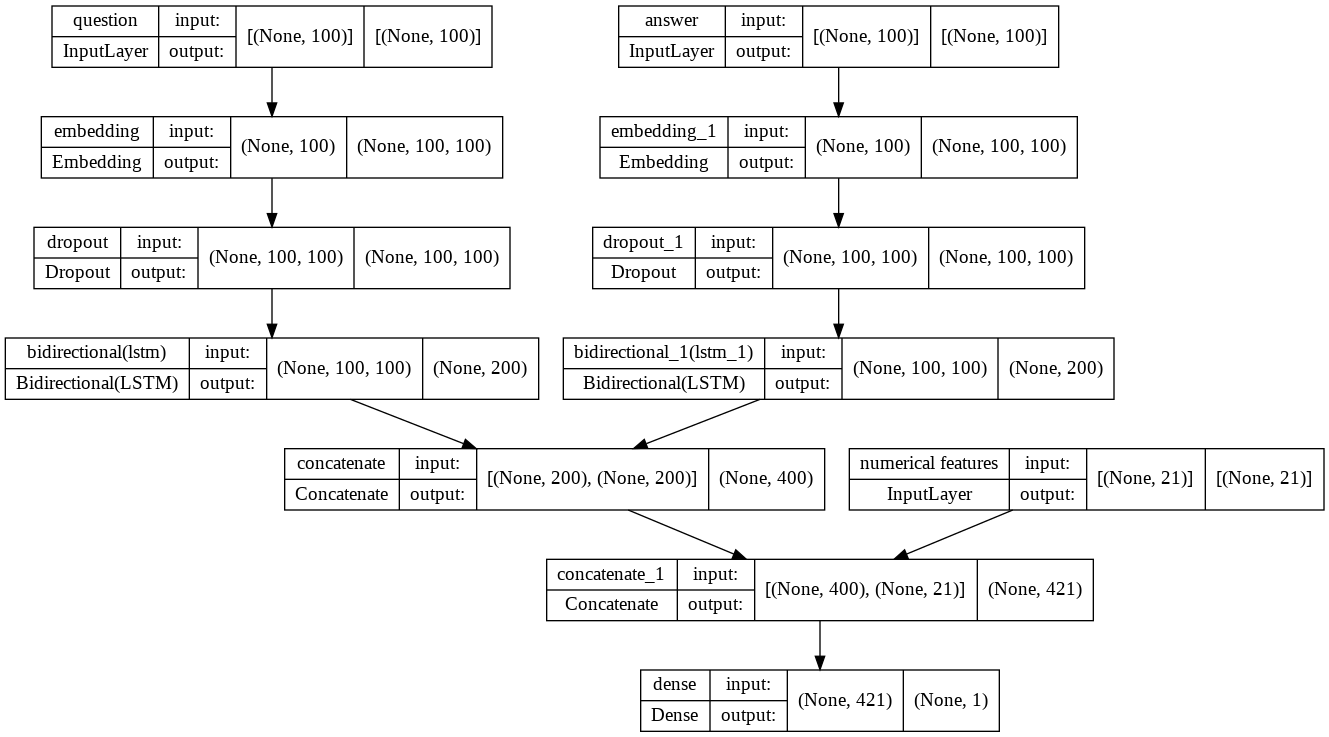}
\end{figure}

Image \ref{img:dl-base-model} shows the Tensorflow plot for the deep learning model we will use, where both question and answer token sequences go through embedding layers before being fed into the LSTM layers to obtain the embedded representations, to finally be concatenated together with the numerical features and fed into the classifier head.

For both of the proposed kind of models (models that only make use of textual features and models that make use of textual and numerical features) we will use the Adam optimizer \cite{https://doi.org/10.48550/arxiv.1412.6980}, and we will sweep across different values for the learning rate parameter, with values comprised between 1e-1 and 1e-6. We will train our models for 5 epochs using a batch size of 512 samples.

\section{Ablation studies}

Once the basic experiments have been carried out we will select the best performing learning rate and we will use it to run a set of ablation studies where we will modify some of the parameters in the original model to check how they affect the model performance.

First we will experiment with the number of sentences found in the question and answer embeddings. We will either not restrict the number of sentences in the textual input, or restrict to a few sentences as maximum. This experiment will allow us to see how dependent the answer ranking performance is on the amount of text input.

Similarly we will also sweep across a few different values for the maximum sequence length accepted by the embedding layers at the model input for both question and answer texts, in order to check how restricting the amount of information input to the model affects its performance.

In another experiment we will use both GloVe embeddings trained on Wikipedia text or Twitter text, in order to find out whether the different training source has an effect on the model performance.

In our experiments we will stack different amounts of LSTMs/Bi-LSTMs layers to increase the model complexity and expressiveness, in order to try to improve the quality of its predictions, as these stacked LSTM layers will allow the model to learn higher order representations that may be useful in our task. We will sweep the amount of stacked LSTM layers from 1 to 4. For these experiments we will set a dropout value in the LSTM layers and in the dropout layer after the embedding layers of 0.25, given that stacking many LSTM layers causes the trainable parameter count to raise by a huge amount and we can introduce overfitting in the training process.

Finally we will try different topologies of classification layers. Hence we will use both single layer and multi layer classification heads. In the case of multi layer classification heads we will experiment with different counts of hidden layers to see whether this has an impact on the model performance, given that we are facing a non linear problem where using a multi-layer classification architecture may be better. We will sweep among the possible settings of no hidden layer, a single hidden layer of size 200, a 2 hidden layer structure with sizes 200 and 100, and finally a 3 hidden layer structure with sizes 200, 100 and 50.
\chapter{Results}

Having defined clearly what our data, models and evaluation methods will be we can now examine the results for each of the experiments described earlier. Initially we will obtain the baseline models' results and then go through each of the proposed experiments that make use of the deep learning models.

\section{Baseline models}

As described in the methods section we have produced different baseline models, making use of different ensembles of features. First we have simple models that only make use of the numerical features. We can see the results for these models in table \ref{results-baseline-num}. We can see that although all models show very similar performance, the xgboost model has better performance across all metrics.

\begin{table}
\centering
\caption{Baseline models results for numerical features}
\label{results-baseline-num}
\begin{tabular}{lrrrr}
\toprule
     classifier &  NDCG@1 &  NDCG@3 &  NDCG@5 &      MRR \\
\midrule
      rf &    0.661982 &    0.861177 &    0.866924 & 0.816129 \\
adaboost &    0.688108 &    0.872320 &    0.877219 & 0.834360 \\
 xgboost &    \textbf{0.691622} &    \textbf{0.874560} &    \textbf{0.879109} & \textbf{0.836995} \\
\bottomrule
\end{tabular}
\end{table}

Afterwards we have created baseline models that make use of a bag-of-words vector space model with TF-IDF weighting. We can see the results for these models in table \ref{results-baseline-text}. Again we observe very similar performance across the different models with xgboost leading across all metrics.

\begin{table}
\centering
\caption{Baseline models results for bag of words features}
\label{results-baseline-text}
\begin{tabular}{lrrrr}
\toprule
     classifier &  NDCG@1 &  NDCG@3 &  NDCG@5 &      MRR \\
\midrule
        rf &    0.496306 &    0.778007 &    0.791852 & 0.712027 \\
  adaboost &    0.493928 &    0.773191 &    0.789303 & 0.692690 \\
   xgboost &    \textbf{0.517005} &    \textbf{0.786364} &    \textbf{0.799849} & \textbf{0.717544} \\
\bottomrule
\end{tabular}
\end{table}

Finally we have created baseline models that make use of the ensemble of numerical features with the bag-of-words vector space model used in the previous experiment. We can see the results for these models in table \ref{results-baseline-ensemble}. We find slightly larger differences between models than on previous tables but again we find that xgboost is the best performing model.

\begin{table}
\centering
\caption{Baseline models results for ensemble of numerical and bag of words features}
\label{results-baseline-ensemble}
\begin{tabular}{lrrrr}
\toprule
     classifier &  NDCG@1 &  NDCG@3 &  NDCG@5 &      MRR \\
\midrule
        rf &    0.600270 &    0.830575 &    0.839080 & 0.774275 \\
  adaboost &    0.641216 &    0.849189 &    0.855798 & 0.805733 \\
   xgboost &    \textbf{0.703243} &    \textbf{0.879077} &    \textbf{0.883278} & \textbf{0.842649} \\
\bottomrule
\end{tabular}
\end{table}

\section{Deep learning models}

We can see the results for the base deep learning model without numerical features on table \ref{results-dl-base-nonum}, and the results for the base deep learning model with numerical features on table \ref{results-dl-base-num}. We see very similar results on both kinds of models, and very slight variation across the learning rate values. As expected adding the numerical features allow the model to perform better than the one using simply textual features.

\begin{table}
\centering
\caption{Deep learning base model results (no numerical features)}
\label{results-dl-base-nonum}
\begin{tabular}{rrrrr}
\toprule
 learning\_rate &   NDCG@1 &   NDCG@3 &   NDCG@5 &      MRR \\
\midrule
      0.100000 & 0.465995 & \textbf{0.775149} & 0.785667 & 0.711209 \\
      0.010000 & 0.455919 & 0.766004 & 0.779665 & 0.703317 \\
      0.001000 & 0.408060 & 0.752389 & 0.761931 & 0.679219 \\
      0.000100 & 0.448363 & 0.764534 & 0.777000 & 0.700882 \\
      0.000010 & \textbf{0.483627} & 0.772991 & \textbf{0.786432} & \textbf{0.713980} \\
      0.000001 & 0.468514 & 0.771911 & 0.782428 & 0.709470 \\
\bottomrule
\end{tabular}
\end{table}

\begin{table}
\centering
\caption{Deep learning base model results (with numerical features)}
\label{results-dl-base-num}
\begin{tabular}{rrrrr}
\toprule
 learning\_rate &   NDCG@1 &   NDCG@3 &   NDCG@5 &      MRR \\
\midrule
      0.100000 & 0.460957 & 0.779047 & 0.784251 & 0.711485 \\
      0.010000 & \textbf{0.496222} & \textbf{0.784116} & \textbf{0.796913} & \textbf{0.726448} \\
      0.001000 & 0.418136 & 0.750740 & 0.764622 & 0.682997 \\
      0.000100 & 0.460957 & 0.780637 & 0.787035 & 0.712720 \\
      0.000010 & 0.410579 & 0.743183 & 0.758903 & 0.676742 \\
      0.000001 & 0.435768 & 0.756588 & 0.770249 & 0.690722 \\
\bottomrule
\end{tabular}
\end{table}

\begin{figure}
\caption{MRR versus answer count}
\label{img:base_model_mrr_n_ans}
\centering
\includegraphics[width=0.5\textwidth]{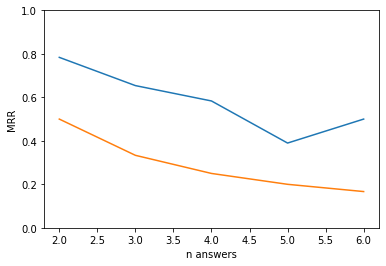}
\end{figure}

\begin{figure}
\caption{MRR histogram}
\label{img:base_model_mrr_hist}
\centering
\includegraphics[width=0.5\textwidth]{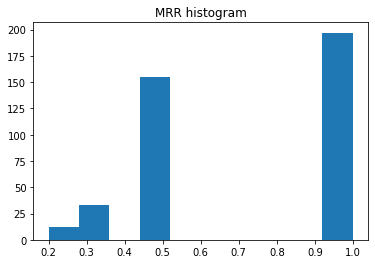}
\end{figure}

In addition to obtaining the NDCG and MRR values for different learning rates we will also plot how the MRR varies with the number of answers posted to each question, as seen on figure \ref{img:base_model_mrr_n_ans}. On this plot we have superposed the curve for the obtained MRR values (blue curve) with the theoretical MRR for a random ranking (orange curve). We will plot a histogram of MRR values so we can see how often the accepted answer is ranked first, as seen in figure \ref{img:base_model_mrr_hist}.

After selecting a learning rate of 0.01 based on the previous results (it maximized the performance of the model with textual and numerical features) we will proceed with the rest of experiments. 

Table \ref{results-dl-max-sent} shows the results for the experiments that varied the maximum sentence count. We see the best performance when there is no constraint for the maximum sentence count, and in general performance degrades as less sentences are considered.

\begin{table}
\centering
\caption{Deep learning maximum sentence count experiments results}
\label{results-dl-max-sent}
\begin{tabular}{rrrrr}
\toprule
 max sentences &   NDCG@1 &   NDCG@3 &   NDCG@5 &      MRR \\
\midrule
            -1 & \textbf{0.526448} & \textbf{0.802948} & \textbf{0.811627} & \textbf{0.746012} \\
             1 & 0.450882 & 0.767712 & 0.780399 & 0.704072 \\
             2 & 0.473552 & 0.777069 & 0.788891 & 0.716649 \\
             3 & 0.496222 & 0.788764 & 0.794942 & 0.725063 \\
             4 & 0.430730 & 0.764053 & 0.772622 & 0.694563 \\
\bottomrule
\end{tabular}
\end{table}

Results for the experiments where the maximum sequence length input to the model is swept while not restricting these sequences to have a maximum sentence count may be found at table \ref{results-dl-max-seq-len}. Even though results are very similar between runs we see that setting a 200 token maximum sequence length performs better across most metrics.

\begin{table}
\centering
\caption{Deep learning maximum sequence length experiments results}
\label{results-dl-max-seq-len}
\begin{tabular}{rrrrr}
\toprule
 max sequence length &   NDCG@1 &   NDCG@3 &   NDCG@5 &      MRR \\
\midrule
                 100 & \textbf{0.486146} & 0.777429 & 0.790337 & 0.717758 \\
                 150 & 0.450882 & 0.775718 & 0.784066 & 0.708690 \\
                 200 & 0.476071 & \textbf{0.783366} & \textbf{0.790849} & \textbf{0.719228} \\
                 250 & 0.460957 & 0.773680 & 0.784307 & 0.710495 \\
                 300 & 0.465995 & 0.774160 & 0.784677 & 0.709950 \\
\bottomrule
\end{tabular}
\end{table}

The next experiment compared the performance of the GloVe embeddings learned from Wikipedia texts with alternate embeddings learned from tweets. Table \ref{results-dl-embedding} shows the results of these experiments. Embeddings learned from Wikipedia content performed slightly better than the Twitter ones.

\begin{table}
\centering
\caption{Deep learning embedding experiments results}
\label{results-dl-embedding}
\begin{tabular}{lrrrr}
\toprule
embedding &   NDCG@1 &   NDCG@3 &   NDCG@5 &      MRR \\
\midrule
  twitter & 0.413098 & 0.755567 & 0.764999 & 0.683291 \\
     wiki & \textbf{0.450882} & \textbf{0.769961} & \textbf{0.781563} & \textbf{0.705542} \\
\bottomrule
\end{tabular}
\end{table}

Our next set of experiments increased the count of stacked LSTM layers after the input embedding. Table \ref{results-dl-lstm} shows the results of these experiments, where we swept the amount of stacked LSTM layers from 1 to 4. We see that having a single LSTM layer performed better across most metrics.

\begin{table}
\centering
\caption{Deep learning LSTM depth experiments results}
\label{results-dl-lstm}
\begin{tabular}{rrrrr}
\toprule
 LSTM depth &   NDCG@1 &   NDCG@3 &   NDCG@5 &      MRR \\
\midrule
          1 & \textbf{0.460957} & 0.769572 & \textbf{0.783233} & \textbf{0.709089} \\
          2 & 0.448363 & 0.764204 & 0.777755 & 0.700672 \\
          3 & 0.455919 & \textbf{0.773680} & 0.781806 & 0.707074 \\
          4 & 0.443325 & 0.765193 & 0.776906 & 0.699370 \\
\bottomrule
\end{tabular}
\end{table}

Finally we experimented how the model behaves when using different hidden layer counts in the classifier head. Table \ref{results-dl-clf-arch} shows the results, where we see that having 2 hidden layers performed better, even though differences seen across runs were small.

\begin{table}
\centering
\caption{Deep learning classifier architecture experiments results}
\label{results-dl-clf-arch}
\begin{tabular}{rrrrr}
\toprule
 hidden layer count &   NDCG@1 &   NDCG@3 &   NDCG@5 &      MRR \\
\midrule
                  0 & 0.405542 & 0.745492 & 0.759374 & 0.677120 \\
                  1 & 0.483627 & 0.784955 & 0.794498 & 0.724139 \\
                  2 & \textbf{0.488665} & 0.\textbf{789663} & \textbf{0.795841} & \textbf{0.727042} \\
                  3 & 0.450882 & 0.774069 & 0.781442 & 0.706447 \\
\bottomrule
\end{tabular}
\end{table}

\chapter{Discussion}

After completing all the proposed experiments we will interpret the results we obtained and draw conclusions from this work.

We will start off by reviewing the results from the different ablation studies we performed on the token sequence generation and the model architecture. As can be seen in the results tables on most experiments all the NDCG and MRR metrics behave in the same direction with minor discrepancies. In case of such discrepancies we will favor MRR when choosing a better result as it considers all the possible answers to a query rather than setting a cut point from which no further answers are considered like NDCG does.

We could clearly see an improvement in all metrics when the numerical features were introduced. It is to note that this improvement was also seen in the baseline models, hence reinforcing our hypothesis that using specific features either originated in the data set source as metadata or extracted manually like we did may improve the results of our answer ranking system.

In terms of the token sequence generation we found better results when not trying to summarize the question and answers content into a reduced amount of sentences. We can easily infer that the model behaves better when seeing more information.

The next experiments tried to increase the sequence length and we find that the results peak with a maximum sequence length of 200 tokens. This is probably due to the fact that this sequence length is enough to contain most of the question or answer content and not requiring padding which may not be meaningful for the model.

After the embedding layer inside the model we find the LSTM layers where representations of the input sequence are generated. Surprisingly we have found that a single LSTM layer outperforms the rest of options. This may be due to the fact that a single LSTM layer is able to produce representations that are meaningful enough for our task, and higher LSTM layer orders only introduce non meaningful higher order representations that do not inform the prediction model any further and may cause the model to over fit.

Finally the experiments where the classifier head structure was changing showed us that using 2 hidden layers improves the model results. This is probably due to the highly non-linear nature of the answer ranking problem, which requires a more complex classifier structure.

Once having reviewed the results of the ablation studies performed before we will compare the performance of the baseline model against the deep learning models we created. If we initially consider the models that do not make use of the numerical features we find that the deep learning model has a similar performance to the best baseline model for bag of words (MRR 0.717544 for xgboost versus MRR 0.713980 for the deep learning model). Then if we consider models that make use of both textual and numerical feature we see that none of the deep learning models variations can even get close to the best baseline model performance (MRR 0.842649 for xgboost versus MRR 0.746012 on the deep learning model maximum sentence count experiment).

We can speculate on a couple of reasons for this poor performance in the deep learning model. First, as the baseline models use a bag of words they are able to observe the contents for the whole of the question or answer text while on the deep learning model the maximum sequence constraint may cause some loss of information. In this case a possible solution would to split the input text into enough input sequences for the model so all of the text may be processed by the model, to later ensemble the prediction results for these split inputs into a single value.

The second possible reason for the lack of performance in the deep learning model may be attributed to the high number of words out of vocabulary when the input embedding stage is performed. As en example this is the report obtained during an embedding pass:

\say{\textit{Converted 16514 words (49474 misses)}}

As we can see the embedding missed many more words than were properly converted. We can look more closely at these embedding misses and observe a sample of 20 words missed during the embedding phase:

\say{\textit{acitvitypackage, belowbody, bindservice, browserpath, callactorsink, createnow, descsortbyname, downloadingprivate, emailbtn, etcby, genfromtxt, hasattribute, hashkey, installthen, inversejoincolumns, justahelperfunction, mockitoannotations, nashornfunction, networksthat, newrandom}}

We can see that all these words are some sort of compound words, formed by two or more words. These are probably variable names from code snippets such as the ones posted by users asking or answering questions related to computer programming problems in Stack Overflow. In this case no possible choice of embedding matrix initialization would help, and we should better be either detecting these compound words and perhaps coding a special token for these situations or detecting whole chunks of code in the question and answer texts. Of course this programming code detection is a whole new problem on its own as computer code may be written in many different languages, with different syntactic rules.
\chapter{Conclusions and future work}

In this work we have looked at the specific problem of ranking multiple answers to questions in community Question Answering sites. This is a very meaningful problem as these sites have experienced big success and hence each of the posted questions is met with multiple answers, of disparate quality and accuracy. It is crucial to the site users and maintainers to be able to sort through all the possible answers to extract meaningful information from them.

We have first compiled a relevant data set encompassing multiple questions and answers from the Stack Overflow site through the Google BigQuery interface. After exploring the data set we have created multiple predictive features to assist with the task of ranking the best answers to each question.

The next step in our work has consisted of creating meaningful baseline models that make use of both the textual content in each question and answer, and of the numerical predictive features we created earlier.

We have then defined multiple metrics to assess the goodness of the models used in the experiments. These metrics such as the NDCG and MRR were created for the information retrieval field where they have been used to great extent with huge success.

Afterwards we have run multiple experiments using a deep learning model structured around embedding and LSTM layers. We have checked the impact of changing the settings in different parts of both the token sequence generation and the model structure.

The outcome of these experiments has been that while leaving out the numerical features we could match the baseline models performance, the deep learning models performance severely lacked the baseline performance when both textual and numerical features were considered.

The existence of many out of vocabulary words on the questions and answers text due to the presence of programming code is the most probable cause for this lack of performance on our deep learning models. A possible solution would be to detect the presence of programming code and tokenize it accordingly.

As future lines of work on this area we could suggest using some of the newest natural language models such as the BERT family of Transformer-based models to try to improve the model prediction capabilities or trying to experiment with the creation of general models that may work across multiple Stack Exchange sub-sites, as on this work we focused on using data from the main Stack Overflow site, focused on computer programming questions. Also as we mentioned the creation of methods to properly detect and delimit the existence of programming code would be a very helpful line of work for the answer ranking problem.

\printbibliography[heading=bibintoc,title={Bibliography}]

\begin{appendices}

\section{Appendix A: SQL queries}
\label{appendix:sql}

SQL query to get the data set is shown below. \texttt{check} will be substituted by = or != to get accepted or not accepted answers, respectively.

\begin{landscape}

\begin{verbatim}
      SELECT 
    question.id as q_id,
    question.Title AS q_title, 
    question.Body AS q_body, 
    question.answer_count as q_answer_count,
    question.accepted_answer_id as q_accepted_a,
    answer.Id AS a_id, 
    answer.Body AS a_body,
    answer.Score as a_score, 
    answer.comment_count AS a_comment_count,
    user.id as user_id,
    user.about_me as user_about,
    user.age as user_age,
    user.creation_date as user_creation_date,
    user.last_access_date as user_last_access_date,
    user.location as user_location,
    user.reputation as user_reputation,
    user.up_votes as user_up_votes,
    user.down_votes as user_down_votes,
    user.views as user_views,
    user.profile_image_url as user_profile_image_url,
    user.website_url as user_website_url,
    CASE WHEN question.accepted_answer_id = answer.Id
      THEN '1'
      ELSE '0' 
    END
    AS a_accepted
  FROM `bigquery-public-data.stackoverflow.posts_answers` AS answer
  JOIN `bigquery-public-data.stackoverflow.posts_questions` question ON question.Id = answer.parent_id
  JOIN `bigquery-public-data.stackoverflow.users` user on user.id = answer.owner_user_id
  WHERE answer.post_type_id = 2 AND question.answer_count > 1 
    AND question.accepted_answer_id IS NOT NULL
    AND question.accepted_answer_id IN (SELECT Id FROM `bigquery-public-data.stackoverflow.posts_answers`)
    AND question.accepted_answer_id {check} answer.Id
     AND EXTRACT(YEAR FROM question.creation_date) = 2016
  ORDER BY question.ID ASC, answer.Id ASC
  LIMIT 100000

\end{verbatim}

\end{landscape}

\end{appendices}

\end{document}